\documentclass{article}

\usepackage{arxiv}

\usepackage[utf8]{inputenc} 
\usepackage[T1]{fontenc}    
\usepackage{hyperref}       
\usepackage{url}            
\usepackage{booktabs}       
\usepackage{amsfonts}       
\usepackage{nicefrac}       
\usepackage{microtype}      
\usepackage{lipsum}
\usepackage{graphicx}
\graphicspath{ {./images/} }

\title{How Does a Multilingual LM Handle Multiple Languages?}

\author{
 Santhosh Kakarla \\
  George Mason University\\
  \texttt{skakarl3@gmu.edu} \\
   \And
 Gautama Shastry Bulusu \\
  George Mason University\\
  \texttt{sbulusuv@gmu.edu} \\
  \And
 Aishwarya Gaddam \\
  George Mason University\\
  \texttt{agaddam3@gmu.edu} \\
  \And
 Maheedhar Sai O M \\
  George Mason University \\
  \texttt{momtrimo@gmu.edu} \\
}

\begin{document}
\maketitle
\begin{abstract}
Multilingual language models (MLMs) have significantly improved due to the quick development of natural language processing (NLP) technologies. These models, such as BLOOM-1.7B, are trained on diverse multilingual datasets and hold the promise of bridging linguistic gaps across languages. However, the extent to which these models effectively capture and utilize linguistic knowledge—particularly for low-resource languages—remains an open research question. This project seeks to critically examine the capabilities of MLMs in handling multiple languages by addressing core challenges in multilingual understanding, semantic representation, and cross-lingual knowledge transfer. \\ \\
   While multilingual language models show promise across diverse linguistic tasks, a notable performance divide exists. These models excel in languages with abundant resources, yet falter when handling less-represented languages. Furthermore, traditional evaluation methods focusing on complex downstream tasks often fail to provide insights into the specific syntactic and semantic features encoded within the models. \\ \\
    This study addresses key limitations in multilingual language models through three primary objectives. First, it evaluates semantic similarity by analyzing whether embeddings of semantically similar words across multiple languages retain consistency, using cosine similarity as a metric. Second, it probes the internal representations of BLOOM-1.7B and Qwen2 through tasks like Named Entity Recognition (NER) and sentence similarity to understand their linguistic structures. Finally, it explores cross-lingual knowledge transfer by examining the models' ability to generalize linguistic knowledge from high-resource languages, such as English, to low-resource languages in tasks like sentiment analysis and text classification. \\ \\
    The results of this study are expected to provide valuable insights into the strengths and limitations of multilingual models, helping to inform strategies for improving their performance. This project aims to deepen our understanding of how MLMs process, represent and transfer knowledge across languages by focusing on a mix of linguistic probing, performance metrics, and visualizations. Ultimately, this study will contribute to advancing language technologies that can effectively support both high- and low-resource languages, thereby promoting inclusivity in NLP applications. 
\end{abstract}


\section{Multilingual Word Embedding Analysis}
The ability to understand relationships between languages is a key challenge in multilingual natural language processing (NLP). In this task, we explore these relationships by analyzing multilingual word embeddings, which are vector representations of words in a shared semantic space. By examining how embeddings of translated words align across languages, we aim to gain insights into the semantic and syntactic similarities between languages. This work is important for several reasons. First, understanding inter-language relationships helps improve machine translation and other multilingual NLP systems. Second, by quantifying these relationships, we can evaluate how well multilingual models like BLOOM-1.7B generalize across languages, especially for underrepresented ones.

\subsection{Methodology}

The task involved multiple steps, beginning with dataset preparation and culminating in the analysis of word embeddings. A corpus of 5,000 English words was extracted from the Gutenberg dataset, a well-established resource for linguistic studies. To enable cross-lingual analysis, translations were generated using Google’s Cloud Translation API, ensuring high-quality multilingual word mappings. Pre-trained transformer-based language models were then employed to generate word embeddings, which encode words in a high-dimensional vector space, capturing both semantic and syntactic relationships. These embeddings served as the foundation for evaluating the semantic consistency of words across languages. \\ \\
Cosine similarity was used as the primary metric to measure the semantic alignment between English embeddings and their translated counterparts. To further explore linguistic relationships, dimensionality reduction techniques such as Principal Component Analysis (PCA) and t-SNE (t-distributed Stochastic Neighbor Embedding) were applied. These techniques enabled the visualization of the high-dimensional embeddings, revealing patterns and clustering that highlighted the proximity and structure of the languages under analysis. The resulting visualizations provided insights into cross-lingual semantic similarities and differences.

\subsection{Results and Analysis}

Google Translate provided high-quality translations, ensuring reliable input for embedding generation. These translations were pivotal in creating accurate multilingual word mappings, which laid the groundwork for subsequent analyses.

Cosine similarity analysis revealed significant insights into the semantic alignment of words across languages. French, Spanish, and German embeddings demonstrated close alignment with English, reflecting shared linguistic roots and similar grammatical structures. These high similarity scores align with the linguistic characteristics of related languages, which share morphological and syntactic features. In contrast, Chinese exhibited substantially lower similarity scores, consistent with its distinct grammatical structures and linguistic features. Figure 1 illustrates the distribution of cosine similarity scores for English and its translated counterparts. French, Spanish, and German showed high scores, whereas Chinese presented significantly lower values, underscoring its linguistic divergence.

\begin{figure}[ht]
\centering
\includegraphics[width=\linewidth]{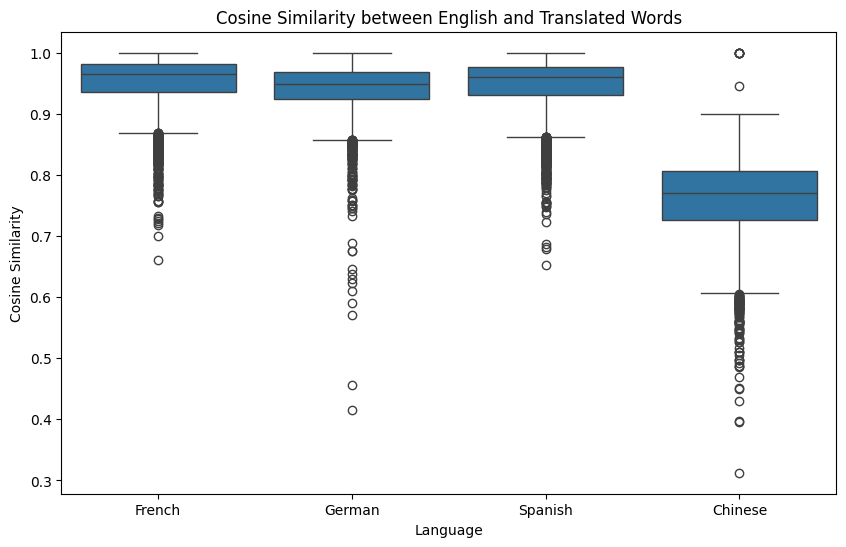}
\caption{Cosine similarity between English and translated words across four languages (French, German, Spanish, and Chinese). The Chinese language exhibits notably lower cosine similarity compared to others.}
\end{figure}

Dimensionality reduction techniques, including Principal Component Analysis (PCA) and t-SNE, were employed to visualize semantic relationships between languages. PCA captured the variance in high-dimensional data, while t-SNE preserved local data structures. The visualizations confirmed clustering patterns among related languages and highlighted semantic distinctions between unrelated languages. As depicted in Figure 2, French, Spanish, and German formed overlapping clusters with English, reflecting their high semantic alignment. Chinese, however, appeared as a distinct cluster, illustrating its linguistic uniqueness and limited overlap with the other languages.

\begin{figure}[ht]
\centering
\includegraphics[width=\linewidth]{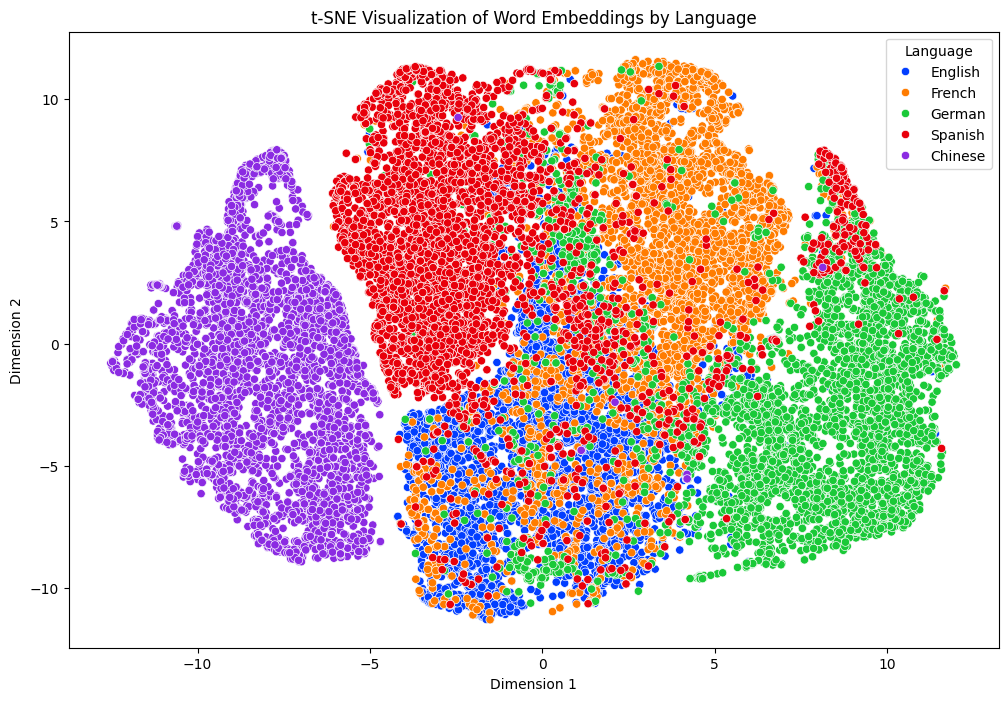}
\caption{t-SNE visualization of word embeddings across five languages: English, French, German, Spanish, and Chinese. Related languages form overlapping clusters, while Chinese forms a distinct, isolated cluster.}
\end{figure}

\subsection{Conclusion}

These results validate the effectiveness of transformer-based word embeddings in capturing semantic and syntactic relationships across languages. The high similarity between English and related languages highlights their shared linguistic features, such as word order and morphological patterns. Conversely, the lower similarity with Chinese underscores its unique linguistic properties. The study demonstrated the potential of transformer-based embeddings in understanding cross-linguistic relationships, supporting the hypothesis that embeddings can effectively represent linguistic proximity and divergence. Future research could extend this analysis to more languages and contextualized embeddings to explore domain-specific applications.

\section{Probing to Understand Model Behavior}

Probing tasks are designed to analyze specific aspects of a model, such as its internal states and behavioral patterns. By probing a model like Bloom-1.7B, researchers can examine its internal workings and learned representations to gain a deeper understanding of how it processes and encodes information. This is achieved by subjecting the model to carefully designed tasks and analyzing its outputs to infer what the model has learned and how this knowledge is distributed across its layers.

Multilingual language models, such as Bloom-1.7B and QWEN-2, are trained to generalize across a variety of languages, which can differ significantly in structure, syntax, and vocabulary. While high-resource languages like English benefit from abundant training data, enabling these models to perform well, low-resource languages like Arabic and Tamil often suffer from limited representation, posing significant challenges. In this context, probing tasks are particularly valuable as they provide insights into how effectively the models transfer linguistic knowledge across different languages.

This study focuses on probing Bloom-1.7B and QWEN-2 through two key tasks. The first task is sentence similarity, which evaluates the models' ability to align semantically similar sentences across languages. Using the Opus parallel corpus, the study measures the similarity of sentence embeddings between English and target languages such as Arabic, Hindi, and Tamil. The second task is named entity recognition (NER), which assesses the models' capability to identify named entities in multilingual data. The CoNLL-2003 dataset is used to evaluate entity recognition performance across both high-resource and low-resource languages.

Probing large models such as BLOOM offers several benefits. First, it provides insights into the internal mechanisms of large language models, enabling researchers to better understand their behavior. This understanding is crucial for designing more interpretable and robust models. Second, probing highlights the areas where the model excels (e.g., syntactic analysis) and areas where it struggles (e.g., handling linguistic ambiguity). Finally, probing improves transparency by offering explanations for the model’s decisions and predictions, which is critical for building trust and improving usability in practical applications.

\subsection{Methodology}

To understand the internal representations of Bloom-1.7B and QWEN-2, hidden state analysis was conducted, focusing on their contributions to both general semantic tasks and task-specific applications. This methodology systematically evaluated the role of each layer in capturing multilingual semantic and syntactic features.

The analysis began with the extraction of hidden states from each layer of Bloom-1.7B and QWEN-2 using input data from the sentence similarity and named entity recognition (NER) tasks. These hidden states were then analyzed layer by layer to determine their relevance to semantic alignment and task-specific performance. For sentence similarity, embeddings were evaluated to identify layers that best preserved semantic consistency across languages. For NER, the study examined how well layers encoded task-specific features like entity boundaries and contextual information.

The process included both quantitative and qualitative evaluations. Quantitatively, metrics such as cosine similarity and F1 scores were used to measure the performance of different layers on the respective tasks. Qualitatively, visualizations such as heat maps and line plots were generated to observe the distribution of hidden state activations across layers. These visualizations provided a clear representation of how linguistic features evolved as input data propagated through the layers.

Key insights emerged from the hidden state analysis. It was observed that the initial layers of both models predominantly captured general semantics, which proved crucial for tasks like sentence similarity. Deeper layers, in contrast, specialized in task-specific features, such as entity recognition in multilingual contexts. For example, layers near the output exhibited heightened sensitivity to named entities and language-specific nuances, underscoring their role in task specialization.

These findings highlight the layered structure and task-specific adaptability of Bloom-1.7B and QWEN-2. The study demonstrates that while general semantics are encoded early in the model, deeper layers play a critical role in addressing the complexities of specific linguistic tasks. This detailed layer-wise understanding not only sheds light on the models' internal architectures but also provides actionable insights for improving performance in multilingual and low-resource language tasks.

\subsection{Results and Analysis}

The performance of the BLOOM model was evaluated by examining sentence similarity across layers for three target languages: Hindi, Tamil, and Arabic. Hindi and Tamil demonstrated remarkably high similarity scores (~0.92-0.95) in the initial layers, suggesting strong initial semantic alignment for these Indo-Aryan and Dravidian languages. In contrast, Arabic began with significantly lower scores (~0.50), indicating early challenges in processing semantic alignment for this Semitic language. As the layers progressed to the mid-range (5-15), Hindi and Tamil showed a gradual decline but maintained relatively high scores (~0.75-0.80), while Arabic exhibited slight improvement, peaking at layer 5 (~0.60). However, a consistent performance gap of ~0.20 persisted between Arabic and Hindi/Tamil. In the deeper layers (15-25), all languages experienced a decline in similarity scores, with Hindi dropping to ~0.68, Tamil decreasing more sharply to ~0.60, and Arabic falling to ~0.48. This pattern suggests a degradation of useful representations in the deeper layers of the model.

\begin{figure}[ht]
\centering
\includegraphics[width=\linewidth]{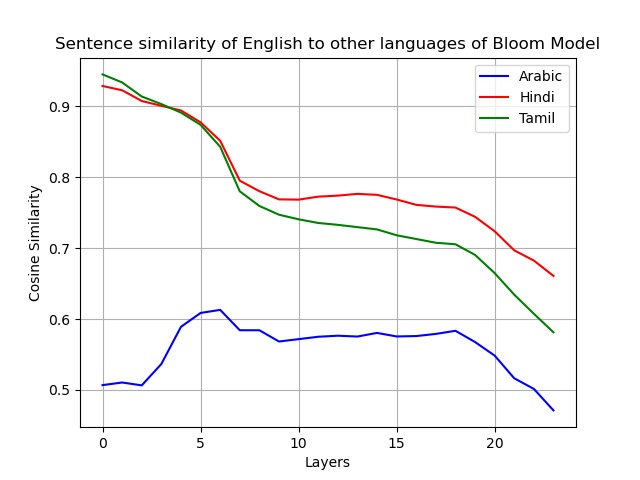}
\caption{Sentence similarity of English to other languages in the BLOOM model. Learning performance decreases as layers increase.}
\end{figure}

In contrast, the QWEN model exhibited distinct cross-lingual alignment capabilities. Hindi and Tamil started with high similarity scores (~0.95) in the initial layers, while Arabic began lower (~0.45) but quickly improved to ~0.80, demonstrating rapid enhancement in cross-lingual alignment. The mid-layers (5-15) showed more stable performance, with Hindi and Tamil hovering around 0.80-0.85 and Arabic stabilizing between 0.70 and 0.75, indicating better preservation of cross-lingual relationships compared to BLOOM. In the deeper layers (15-25), performance degradation was minimal. Hindi and Tamil maintained scores around 0.85-0.90, while Arabic showed slight improvement, reaching ~0.80, suggesting that QWEN’s architecture is more resilient in preserving useful representations.

\begin{figure}[ht]
\centering
\includegraphics[width=\linewidth]{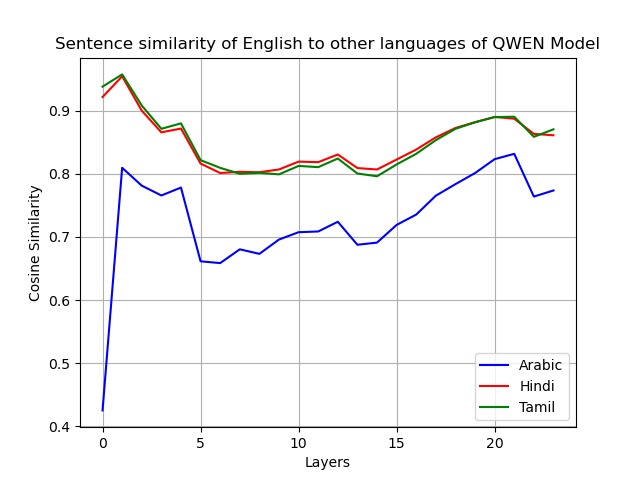}
\caption{Sentence similarity of English to other languages in the QWEN model. Arabic shows improved performance compared to the BLOOM model.}
\end{figure}

The hidden state analysis further revealed critical insights into the internal dynamics of the models. In the early layers (0-5), hidden state values remained near zero with minimal variation, indicating stable initial processing and effective basic feature extraction. As the data propagated through the mid-layers (5-20), a slight positive trend in hidden state values was observed, stabilizing around 0.05-0.10, reflecting consistent and effective information processing. However, the deeper layers (20-25) exhibited a sharp decline in hidden state values, dropping to -1.0 at layer 25, indicating severe degradation of useful representations. This suggests potential optimization challenges in the deeper layers, particularly in BLOOM.

\begin{figure}[ht]
\centering
\includegraphics[width=\linewidth]{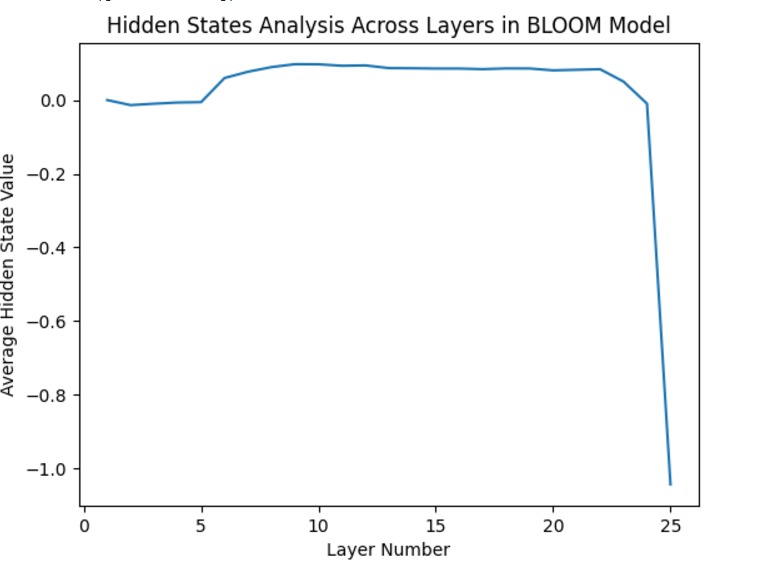}
\caption{Hidden state analysis across layers in the BLOOM model. The hidden state value decreases sharply in the deeper layers, indicating degradation of useful representations.}
\end{figure}

These analyses provide a comprehensive understanding of the BLOOM and QWEN models’ internal behavior, highlighting areas of strength and potential improvement for multilingual and low-resource language tasks.

\subsection{Conclusions and Future Insights}

This study has provided valuable insights into the performance and internal dynamics of the BLOOM and QWEN models. The findings demonstrate that QWEN offers superior cross-lingual alignment, particularly for low-resource languages like Arabic, while maintaining more stable performance across layers. BLOOM, on the other hand, shows significant degradation in deeper layers, highlighting architectural challenges in maintaining useful representations for low-resource languages. These results emphasize the importance of robust architectural designs in handling diverse linguistic features.

Future work should focus on improving cross-lingual alignment for low-resource languages through methods like data augmentation, self-supervised pretraining, or leveraging synthetic data to enhance semantic consistency. Further architectural optimization, such as dynamic layer-wise learning rates and advanced attention mechanisms, could help mitigate deeper layer degradation, particularly in BLOOM. Expanding the scope of probing tasks to include syntactic parsing, coreference resolution, and sentiment analysis will offer a more comprehensive evaluation of model capabilities. Moreover, analyzing model behavior in real-world applications, such as dialogue systems or multimodal contexts, could deepen our understanding of their strengths and limitations. Fine-tuning for domain-specific tasks, such as healthcare or legal text analysis, would also provide practical insights for optimizing these models for specialized applications, ultimately contributing to more robust, interpretable, and efficient multilingual systems.

\section{Cross-Lingual Transferability}

The ability to transfer linguistic knowledge across languages is a cornerstone of multilingual natural language processing (NLP). With the rapid advancement of transformer-based models, cross-lingual transferability has emerged as a vital area of research to address the disparity in linguistic resources between high-resource and low-resource languages. High-resource languages like English benefit from extensive datasets and pretraining resources, enabling the development of robust NLP systems. In contrast, low-resource languages, such as Arabic and Swahili, lack sufficient linguistic data, which limits the performance and accessibility of NLP tools in these languages. Addressing this resource imbalance is crucial for building equitable and inclusive NLP systems that serve diverse linguistic communities.

This study explores the cross-lingual capabilities of two state-of-the-art multilingual models—BLOOM-560m and BERT-base Multilingual Cased. By evaluating their performance on multilingual tasks, this research investigates their ability to generalize knowledge learned from high-resource languages, such as English, and transfer it to low-resource languages like Arabic and Swahili. The models' cross-lingual transferability is assessed through tasks such as text classification and named entity recognition (NER), using metrics like accuracy and F1 score to quantify their effectiveness. While BERT-base Multilingual Cased has been widely adopted in various multilingual tasks, BLOOM-560m represents a newer architecture with the potential to outperform existing models by leveraging advanced pretraining techniques and a larger training corpus.

Building upon prior studies that have examined the cross-lingual capabilities of models like XLM-R, this work aims to provide deeper insights into the strengths and limitations of modern multilingual architectures. By identifying areas for improvement and optimization, the study contributes to the development of more robust and adaptable NLP systems that can effectively bridge the gap between high-resource and low-resource languages. The findings from this research hold significant implications for the broader goal of advancing equitable and inclusive language technologies across diverse linguistic contexts.

\subsection{Methodology}

This study investigates cross-lingual transferability by leveraging two prominent multilingual models, BLOOM-560m and BERT-base Multilingual Cased. These models are selected for their distinct architectures and established roles in multilingual NLP tasks. The XNLI dataset, a widely recognized benchmark for natural language inference tasks, serves as the primary data source. It provides aligned examples across multiple languages, making it well-suited for evaluating the transferability of linguistic knowledge from high-resource to low-resource languages.

The study begins by fine-tuning the models on English data from the XNLI training set, ensuring that both models are exposed to high-resource language data. After fine-tuning, the models are evaluated on validation datasets for English, Arabic, and Swahili to determine their cross-lingual transfer performance. This evaluation process highlights the models' ability to generalize linguistic knowledge acquired from English and apply it effectively to low-resource languages. Tokenization is performed using the respective tokenizers for BLOOM-560m and BERT-base Multilingual Cased, and input sequences consisting of premise and hypothesis pairs are truncated or padded to a fixed length of 128 tokens to maintain consistency across all experiments.

\subsection{Results and analysis}

The results demonstrate that BLOOM-560m consistently outperforms BERT-base Multilingual Cased in terms of accuracy, F1 score, and transfer efficiency across all three evaluated languages: English, Arabic, and Swahili. Figure 6 presents a comparison of accuracy and F1 scores, where BLOOM-560m achieves superior performance across all languages. Specifically, BLOOM-560m achieves an accuracy of 0.70, 0.59, and 0.50 for English, Arabic, and Swahili, respectively, while BERT-base Multilingual Cased achieves 0.54, 0.47, and 0.40. Similarly, BLOOM-560m maintains higher F1 scores, with 0.70, 0.49, and 0.46 compared to BERT-base Multilingual Cased's 0.54, 0.46, and 0.30.

Transfer efficiency, which measures relative performance compared to English as a baseline, is illustrated in Figures 7 and 8. BERT-base Multilingual Cased achieves transfer efficiency scores of 0.84 and 0.71 for Arabic and Swahili, respectively, while BLOOM-560m outperforms it with 0.87 and 0.73. These findings emphasize BLOOM-560m's ability to retain and apply linguistic knowledge more effectively across diverse languages, highlighting its architectural advantages.

Both models exhibit their best performance in English, underscoring the influence of high-quality and abundant training data. However, the notable drop in performance for Arabic and Swahili reveals the persistent challenges of low-resource language processing. BLOOM-560m's relatively stable performance in these settings positions it as a strong candidate for further development in multilingual and low-resource language applications.

These results underscore the architectural strengths of BLOOM-560m and its potential for advancing equitable NLP technologies. Figures 6, 7, and 8 provide a comprehensive visual representation of the models' comparative performance across these metrics.

\begin{figure}[ht]
\centering
\includegraphics[width=\linewidth]{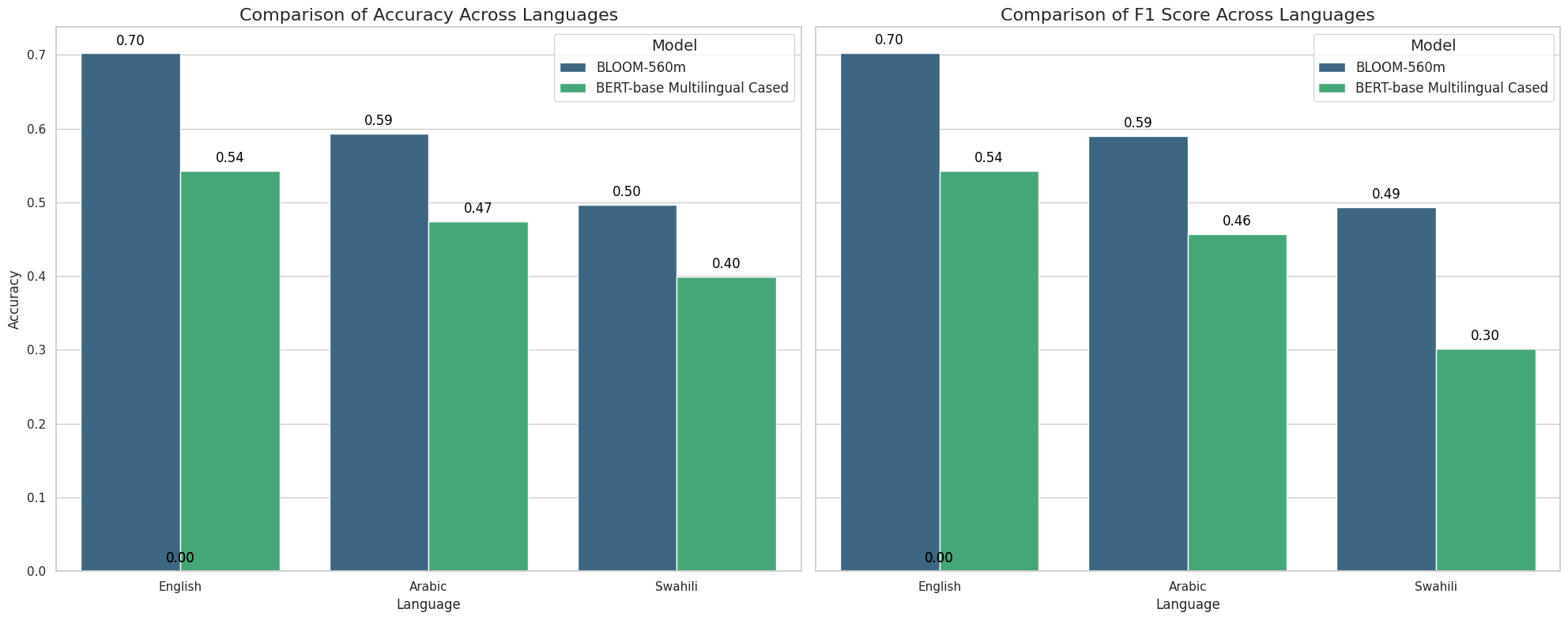}
\caption{Comparison of Accuracy and F1 Score Across Languages for BLOOM-560m and BERT-base Multilingual Cased.}
\end{figure}

\begin{figure}[ht]
\centering
\includegraphics[width=\linewidth]{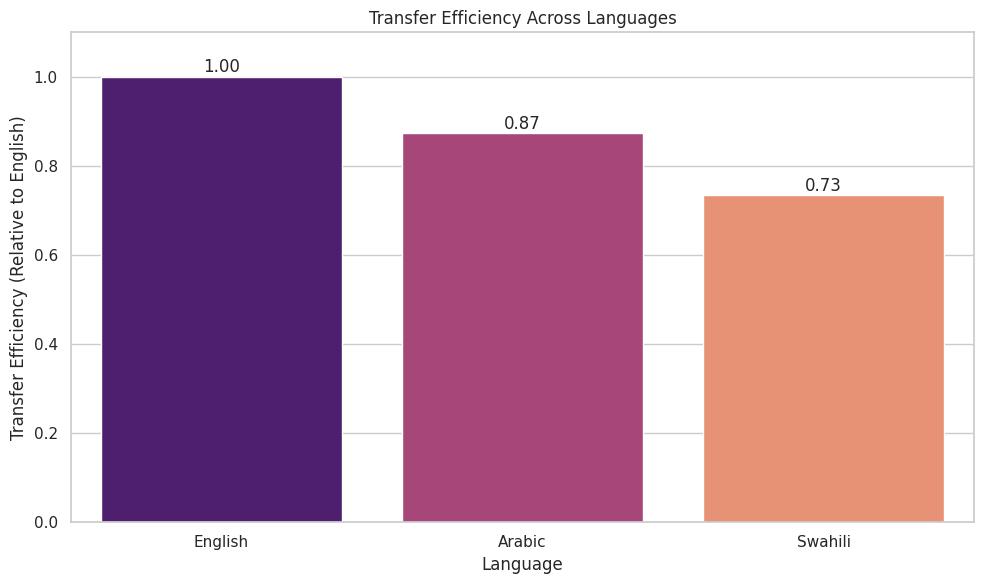}
\caption{Transfer Efficiency Across Languages for BERT-base Multilingual Cased.}
\end{figure}

\begin{figure}[ht]
\centering
\includegraphics[width=\linewidth]{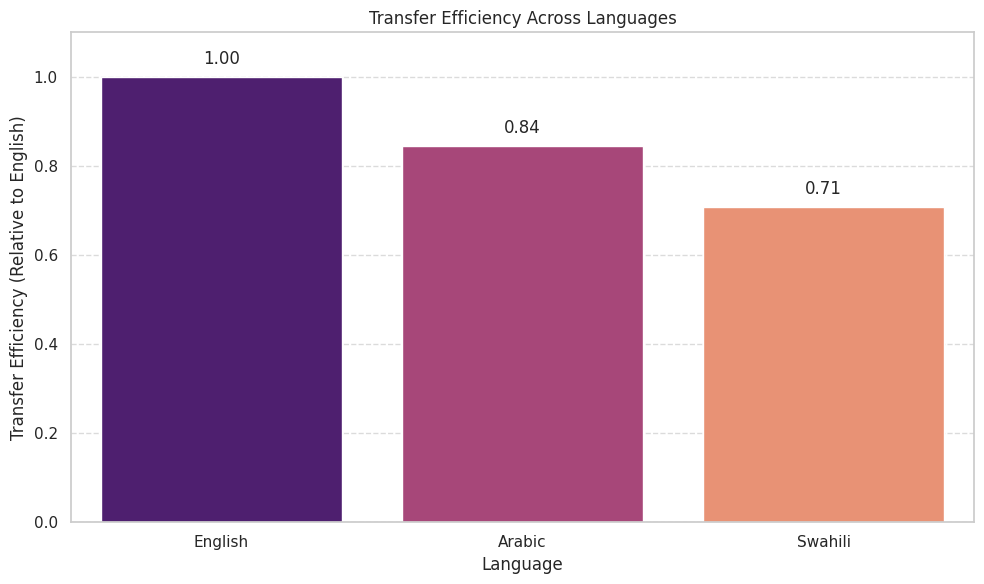}
\caption{Transfer Efficiency Across Languages for BLOOM-560m.}
\end{figure}

These findings highlight the potential of BLOOM-560m in advancing NLP for diverse linguistic contexts and provide a foundation for future research focused on addressing the unique challenges of low-resource languages.

\subsection{Conclusion and Future Insights}

BLOOM-560m demonstrates significant potential for cross-lingual tasks, consistently outperforming BERT-base Multilingual Cased in accuracy and F1 scores across all evaluated languages. However, the results also highlight a persistent gap in transferability to low-resource languages, with both models showing their weakest performance on Swahili. This suggests that more advanced techniques are required to fully bridge the gap between high-resource and low-resource languages. While BLOOM-560m exhibits better overall performance, its transfer efficiency for low-resource languages like Swahili may still be slightly lower than expected in certain scenarios, indicating room for architectural optimization.

Future work should focus on several key areas to address these challenges. First, incorporating more diverse multilingual datasets, including those with code-switched text and data from underrepresented domains, could enhance the models' ability to generalize. Second, exploring advanced fine-tuning techniques such as few-shot learning, meta-learning, and multi-task learning could improve model adaptation in low-resource scenarios. Additionally, integrating dynamic masking strategies and domain-specific pretraining might further enhance cross-lingual transferability. These approaches collectively aim to advance the development of more robust, adaptable, and equitable multilingual NLP systems.

\section{Limitations}
The key limitations of this study include the restricted training on a limited set of languages due to GPU resource constraints, which limits the generalizability of findings across diverse linguistic families. Additionally, while the models perform well for high-resource languages, their effectiveness for low-resource languages remains suboptimal, highlighting the need for better representation learning.

\bibliographystyle{unsrt}  


\end{document}